%% file: iclr2020_conference.tex
\documentclass{article} 
\usepackage{iclr2020_conference,times}

\input{math_commands.tex}

\usepackage{hyperref}
\usepackage{url}
\usepackage{graphicx}
\usepackage{booktabs}
\definecolor{ao}{rgb}{0.0, 0.5, 0.0}

\title{Improving Yor{\`u}b{\'a} Diacritic Restoration}

\author{Iroro Fred \d{\`O}n\d{\`o}m\d{\`e} Orife \\
Niger-Volta LTI\\
\And
David I. Ad{\'e}lan{\'i} \\
Saarland University, Niger-Volta LTI \\
\And
Timi Fasubaa \\
Niger-Volta LTI \\
\And
Victor Williamson \\
University of Wisconsin-Milwaukee \\
\And
Wuraola Fisayo Oyewusi\\
Data Science Nigeria \\
\And
\d{O}l{\'a}mil{\'e}kan Wahab \\
Niger-Volta LTI \\
\And
K\d{\'{o}}l\'{a} T\'{u}b\d{\`{o}}s\'{u}n \\
Yor{\`u}b{\'a} Name \\
\\
\texttt{\footnotesize iroro@alumni.cmu.edu, didelani@lsv.uni-saarland.de} \vspace*{-.2em} \\
\texttt{\footnotesize timifasubaa@berkeley.edu,  victorlamont05@gmail.com} \vspace*{-.2em} \\
\texttt{\footnotesize oyewusiwuraola@gmail.com, olamyy53@gmail.com, kolatubosun@gmail.com}
}

\iclrfinalcopy 
\begin{document}

\maketitle

\section{Introduction}\label{sec:introduction}

Yor{\`u}b{\'a} is a tonal language spoken by more than 40 Million people in the countries of Nigeria, Benin and Togo in West Africa. The phonology is comprised of eighteen consonants, seven oral vowel and five nasal vowel phonemes with three kinds of tones realized on all vowels and syllabic nasal consonants \citep{akinlabi2004sound}. Yor{\`u}b{\'a} orthography makes notable use of tonal diacritics, known as \emph{am{\'i} oh{\`u}n}, to designate tonal patterns, and orthographic diacritics like underdots for various language sounds \citep{adegbola2012quantifying, wells2000orthographic}.

Diacritics provide morphological information, are crucial for lexical disambiguation and pronunciation, and are vital for any computational Speech or Natural Language Processing (NLP) task. To build a robust ecosystem of \emph{Yor{\`u}b{\'a}-first} language technologies, Yor{\`u}b{\'a} text must be correctly represented in computing environments. The ultimate objective of automatic diacritic restoration (ADR) systems is to facilitate text entry and text correction that encourages the correct orthography and promotes quotidian usage of the language in electronic media.

\subsection{Ambiguity in non-diacritized text}
The main challenge in non-diacritized text is that it is very ambiguous \citep{orife2018adr, asahiah2017restoring, adegbola2012quantifying, de2007automatic}. ADR attempts to decode the ambiguity present in undiacritized text. Adegbola et al. assert that for ADR the ``prevailing error factor is the number of valid alternative arrangements of the diacritical marks that can be applied to the vowels and syllabic nasals within the words" \citep{adegbola2012quantifying}.

\begin{table}[h]
\caption{Diacritic characters with their non-diacritic forms}
\label{ambiguity-table}
\begin{center}
  \begin{tabular}{lcl}
    \multicolumn{2}{c}{\bf Characters} & \textbf{Examples}  \\
	\toprule
    {\`a} {\'a} \v{a} & \textbf{a} & gb{\`a} \emph{(spread)}, gba \emph{(accept)}, gb{\'a} \emph{(hit)}    \\  
    {\`e} {\'e} \d{e} \d{\`e} \d{\'e} & \textbf{e} & {\`e}s{\`e} \emph{(dye)}, \d{e}s\d{\`e} \emph{(foot)}, es{\'e} \emph{(cat)}\\
    {\`i} {\'i} & \textbf{i} & {\`i}l{\`u} \emph{(drum)}, ilu \emph{(opener)}, {\`i}l{\'u} \emph{(town)}\\  
    {\`o} {\'o} \d{o} \d{\`o} \d{\'o} \v{o} & \textbf{o} & ar\d{o} \emph{(an invalid)}, ar{\'o} \emph{(indigo)}, {\`a}r{\`o} \emph{(hearth)}, {\`a}r\d{o} \emph{(funnel)}, {\`a}r\d{\`o} \emph{(catfish)}\\  
    {\`u} {\'u} \v{u} & \textbf{u} & k{\`u}n \emph{(to paint)}, kun \emph{(to carve)}, k{\'u}n \emph{(be full)} \\ \\
	\midrule
    {\`n} {\'n} \={n} & \textbf{n} & {\`n} (a negator), {n} \emph{(I)}, {\'n} (continuous aspect marker) \\  
    \d{s} & \textbf{s} &  \d{s}{\`a} \emph{(to choose)}, \d{s}{\'a} \emph{(fade)}, {s}{\`a} \emph{(to baptise)}, {s}{\'a} \emph{(to run)} \\
  \bottomrule
  \end{tabular}
\end{center}
\end{table}

\subsection{Improving generalization performance}

To make the first open-sourced ADR models available to a wider audience, we tested extensively on colloquial and conversational text. These soft-attention seq2seq models \citep{orife2018adr}, trained on the first three sources in Table \ref{tab:training_datasets}, suffered from domain-mismatch generalization errors and appeared particularly weak when presented with contractions, loan words or variants of common phrases. Because they were trained on majority Biblical text, we attributed these errors to low-diversity of sources and an insufficient number of training examples. To remedy this problem, we aggregated text from a variety of online public-domain sources as well as actual books. After scanning physical books from personal libraries, we successfully employed commercial Optical Character Recognition (OCR) software to concurrently use English, Romanian and Vietnamese characters, forming an \emph{approximative superset} of the Yor{\`u}b{\'a} character set. Text with inconsistent quality was put into a special queue for subsequent human supervision and manual correction. The post-OCR correction of H{\'a}{\`a} {\`E}n{\`i}y{\`a}n, a work of fiction of some 20,038 words, took a single expert two weeks of part-time work by to review and correct. Overall, the new data sources comprised varied text from conversational, various literary and religious sources as well as news magazines, a book of proverbs and a Human Rights declaration.

\section{Methodology}\label{sec:methods}

\paragraph{Experimental setup}\label{sec:experimental}
Data preprocessing, parallel text preparation and training hyper-parameters are the same as in \citep{orife2018adr}. Experiments included evaluations of the effect of the various texts, notably for JW300, which is a disproportionately large contributor to the dataset. We also evaluated models trained with pre-trained FastText embeddings to understand the boost in performance possible with word embeddings \citep{alabi2019massive, bojanowski2017enriching}. Our training hardware configuration was an AWS EC2 p3.2xlarge instance with \texttt{OpenNMT-py} \citep{opennmt}.

 \begin{table}[h]
  \caption{Data sources, prevalence and category of text}
  \label{tab:training_datasets}
  \begin{center}
  \begin{tabular}{rll}
    \toprule
    \textbf{\# words} & \textbf{Source or URL}  & \textbf{Description} \\
    \midrule
    24,868 & rma.nwu.ac.za  & Lagos-NWU corpus \\  
    50,202 & theyorubablog.com & language blog \\ 
    910,401 & bible.com/versions/911 & Biblica (NIV) \\
    \midrule
    11,488,825 & opus.nlpl.eu & JW300 \\
    831,820 & bible.com/versions/207 & Bible Society Nigeria (KJV)\\
    177,675 & GitHub & Embeddings dataset (mixed) \\
    142,991 & GitHub & Language ID corpus \\
    47,195 &  & Yor{\`u}b{\'a} lexicon \\
    29,338 & yoruba.unl.edu & Proverbs \\
    2,887 & unicode.org/udhr & Human rights edict \\
    \midrule
    150,360 & Private sources & Conversational interviews \\
    15,281 & Private sources & Short stories \\
    20,038 & OCR & H{\'a}{\`a} {\`E}n{\`i}y{\`a}n (Fiction) \\
    \midrule
    \midrule
    28,308 & yo.globalvoices.org & Global Voices news \\
    \bottomrule
  \end{tabular} 
  \end{center}
\end{table}

\paragraph{A new, modern multi-purpose evaluation dataset}\label{sec:evaldataset} To make ADR productive for users, our research experiments needed to be guided by a test set based around modern, colloquial and not exclusively literary text. After much review, we selected Global Voices, a corpus of journalistic news text 
from a multilingual community of journalists, translators, bloggers, academics and human rights activists \citep{global_voices}.

\section{Results}\label{sec:results}
We evaluated the ADR models by computing a single-reference BLEU score using the Moses \texttt{multi-bleu.perl} scoring script, the predicted perplexity of the model's own predictions and the Word Error Rate (WER). All models with additional data improved over the 3-corpus soft-attention baseline, with JW300 providing a \{33\%, 11\%\}  boost in BLEU and absolute WER respectively. Error analyses revealed that the Transformer was robust to receiving digits, rare or code-switched words as input and degraded ADR performance gracefully. In many cases, this meant the model predicted the undiacritized word form or a related word from the context, but continued to correctly predict subsequent words in the sequence. The FastText embedding provided a small boost in performance for the Transformer, but was mixed across metrics for the soft-attention models. 
 \begin{table}[h]
  \caption{BLEU, predicted perplexity \& WER on the Global Voices testset}
  \label{tab:results-appendix}
 \begin{center}
  \begin{tabular}{lccc}
    \toprule
    \textbf{Model} & \textbf{BLEU} &\textbf{Perplexity} &\textbf{WER\%}\\
    \bottomrule
    \\
    Soft-attention model from \citep{orife2018adr} & 26.53 & 1.34 & 58.17 \\
    \midrule
	\hspace{5mm} + Language ID corpus & 42.52 & 1.69 & 33.03 \\ 
	\hspace{10mm} ++ Interview text & 42.23 & 1.59 & 32.58 \\
	\hspace{5mm} + All new text \it{minus} JW300 & 43.39 & 1.60 & 31.87 \\ 
	\hspace{5mm} + All new text & \textbf{59.55} & 1.44 & \textbf{20.40}\\ 
	\hspace{5mm} + All new text with FastText embedding & 58.87 & \textbf{1.39} & 21.33 \\ 
    \midrule
	Transformer model \\
	 \hspace{5mm} + All new text \it{minus} JW300 & 45.68 & 1.95 & 34.40\\
	 \hspace{5mm} + All new text & 59.05 & \textbf{1.40} &  23.10\\
	 \hspace{5mm} + All new text + FastText embedding & \textbf{59.80} & 1.43 & \textbf{22.42}\\ 
    \bottomrule
  \end{tabular}
  \end{center}
\end{table}

\section{Conclusions and Future Work}

Promising next steps include further automation of our human-in-the-middle data-cleaning tools, further research on contextualized word embeddings for Yor{\`u}b{\'a} and serving or deploying the improved ADR models\footnote{\url{https://github.com/Niger-Volta-LTI/yoruba-adr}}\footnote{\url{https://github.com/Niger-Volta-LTI/yoruba-text}} in user-facing applications and devices.  

\bibliography{iclr2020_conference}
\bibliographystyle{iclr2020_conference}

\appendix
\section{Appendix}\label{sec:appendix}

\begin{table}[h]
\caption{The best performing Transformer model trained with the FastText embedding was used to generate predictions. The Baseline model is the 3-corpus soft-attention model. ADR errors are in \textbf{\textcolor{red}{red}}, robust predictions of rare, loan words or digits in \textbf{\textcolor{ao}{green}}. }\label{results}
\begin{center}
  \begin{tabular}{ll}
  	\toprule \\
	   \textbf{Source:}  & mo ro o wipe awon obirin ti o ronu lati se ise ti okunrin maa n se gbodo gberaga .\\
       \textbf{Reference:} &  mo r{\`o} {\'o} w{\'i}p{\'e} {\`a}w\d{o}n ob{\`i}rin t{\'i} {\'o} ron{\'u} l{\'a}ti \d{s}e i\d{s}\d{{\'e}} t{\'i} \d{o}k{\`u}nrin m{\'a}a {\'n} \d{s}e gb\d{o}d\d{{\`o}} gb{\'e}raga . \\
	   \textbf{Prediction:} &   mo r{\`o} {\'o} w{\'i}p{\'e} {\`a}w\d{o}n ob{\`i}rin t{\'i} {\'o} ron{\'u} l{\'a}ti \d{s}e i\d{s}\d{{\'e}} t{\'i} \d{o}k{\`u}nrin m{\'a}a {\'n} \d{s}e gb\d{o}d\d{{\`o}} gb{\'e}raga . \\ 
	   \textbf{Baseline:} &   mo r{\`o} {\'o} w{\'i}p{\'e} {\`a}w\d{o}n ob{\`i}rin t{\'i} {\'o} ron{\'u} l{\'a}ti \d{s}e i\d{s}\d{{\'e}} t{\'i} \d{o}k{\`u}nrin m{\'a}a \textbf{\textcolor{red}{dar{\'i} s\d{{\`o}}r\d{{\`o}} l{\'a}\d{s}e l\d{{\'o}}kan}} . \\ \\

	\midrule \\ 
		\textbf{Source:}     & bi o tile je pe egbeegberun ti pada sile .  \\
		\textbf{Reference:}  & b{\'i} {\'o} til\d{{\`e}}  j\d{{\'e}}  p{\'e} \d{e}gb\d{e}\d{e}gb\d{{\`e}}r{\'u}n ti pad{\`a} s{\'i}l{\'e} . \\
		\textbf{Prediction:} & b{\'i} {\'o} til\d{{\`e}}  j\d{{\'e}}  p{\'e} \d{e}gb\d{e}\d{e}gb\d{{\`e}}r{\'u}n ti pad{\`a} s{\'i}l{\'e} .\\ 
		\textbf{Baseline:} & b{\'i} {\'o} til\d{{\`e}}  j\d{{\'e}}  p{\'e} \d{e}gb\d{e}\d{e}gb\d{{\`e}}r{\'u}n t{\'i} pad{\`a} \textbf{\textcolor{red}{s{\'i}l\d{{\`e}} s\d{{\`o}}r\d{{\`o}}}}\\ \\
    
    \midrule \\ 
    	\textbf{Source:} &       mo awon ondije si ipo aare naijiria odun \textbf{\textcolor{ao}{2019}} \\
		\textbf{Reference:}  &   m\d{o} {\`a}w\d{o}n {\`o}{\`n}d{\'i}je s{\'i} ip{\`o} {\`a}{\`a}r\d{e} n{\`a}{\`i}j{\'i}r{\'i}{\`a} \d{o}d{\'u}n \textbf{\textcolor{ao}{2019}} \\
		\textbf{Prediction:} &   m\d{o} {\`a}w\d{o}n \textbf{\textcolor{red}{ondije}} s{\'i} ip{\`o} {\`a}{\`a}r\d{e}  n{\`a}{\`i}j{\'i}r{\'i}{\`a} \d{o}d{\'u}n \textbf{\textcolor{ao}{2019}} \\ 
		\textbf{Baseline:} &  mo {\`a}w\d{o}n \textbf{\textcolor{red}{ojoj{\'u}m\d{{\'o}}}} s{\'i} ip{\`o} \textbf{\textcolor{red}{{\`a}{\'a}r\d{{\`e}}}} n{\`a}\textbf{\textcolor{red}{{\'i}}}j{\'i}r{\'i}{\`a} \d{o}d{\'u}n \textbf{\textcolor{red}{kiki}} \\\\	 

	\midrule \\
   	\textbf{Source:}     & iriri akobuloogu \textbf{\textcolor{ao}{zone9}} ilu ethiopia je apeere . \\
	\textbf{Reference:}  & {\`i}r{\'i}r{\'i} \textbf{ak\d{o}b{\'u}l\d{{\'o}}\d{{\`o}}g{\`u}} \textbf{\textcolor{ao}{zone9}}  {\`i}l{\'u} ethiopia j\d{{\'e}} {\`a}p\d{e}\d{e}r\d{e} . \\
	\textbf{Prediction:} & {\`i}r{\'i}r{\'i} \textbf{\textcolor{red}{akobuloogu}} \textbf{\textcolor{ao}{or{\'i}l{\`e}}} {\`i}l{\'u} ethiopia j\d{{\'e}} {\`a}p\d{e}\d{e}r\d{e} . \\ 
	\textbf{Baseline:} & {\`i}r{\'i}r{\'i} \textbf{\textcolor{red}{{\`a}w\d{o}n}} {\`i}l{\'u} \textbf{\textcolor{red}{esinsin ar{\'a}k{\`u}nrin}} j\d{{\'e}} {\`a}p\d{e}\d{e}r\d{e} .\\ \\
  	
    \midrule \\
	\textbf{Source:}     & alaga akoko ilu-ti-ko-fi-oba-je ti china mao zedong ti yo awon eniyan ninu isoro . \\
	\textbf{Reference:}  & al{\'a}ga {\`a}k\d{{\'o}}k\d{{\'o}} \textbf{{\`i}l{\'u}-t{\'i}-k{\`o}-fi-\d{o}ba-j\d{e}} ti \textbf{\textcolor{ao}{china mao zedong}} ti y\d{o} {\`a}w\d{o}n {\`e}n{\`i}y{\`a}n n{\'i}n{\'u} {\`i}\d{s}{\`o}ro . \\
	\textbf{Prediction:} & al{\'a}ga {\`a}k\textbf{\textcolor{red}{{\'o}k{\`o} ilu-ti-ko-fi-oba-je}} ti \textbf{\textcolor{ao}{china mao tse}} ti y\d{o} {\`a}w\d{o}n {\`e}n{\`i}y{\`a}n n{\'i}n{\'u} {\`i}\d{s}{\`o}ro . \\ 
	\textbf{Baseline:} & \textbf{\textcolor{red}{j{\'e}h{\'o}\d{s}{\'a}f{\'a}t{\`i} {\`a}k{\'o}k{\`o} sam{\'a}r{\'i}{\`a}}} t{\'i} china \textbf{\textcolor{red}{l\d{e}\d{s}\d{e}k\d{e}\d{s}\d{e} ap{\'a} t{\'i} w{\`a} at{\'i}}} {\`e}niy{\`a}n n{\'i}n{\'u} {\`i}\d{s}{\`o}ro . \\ \\
    \bottomrule
  \end{tabular}
\end{center}
\end{table} 

\end{document}

%% file: math_commands.tex
\usepackage{amsmath,amsfonts,bm}

\def\eqref#1{equation~\ref{#1}}

\def\1{\bm{1}}

\DeclareMathAlphabet{\mathsfit}{\encodingdefault}{\sfdefault}{m}{sl}
\SetMathAlphabet{\mathsfit}{bold}{\encodingdefault}{\sfdefault}{bx}{n}

